\documentclass[letterpaper, 10 pt, conference]{ieeeconf}  % Comment this line out if you need a4paper

\IEEEoverridecommandlockouts                              % This command is only needed if 
\usepackage{multirow}
\usepackage{graphics} % for pdf, bitmapped graphics files
\usepackage{graphicx}
\usepackage{subcaption}
\usepackage[dvipsnames, table]{xcolor}
\usepackage{amsmath} % assumes amsmath package installed
\usepackage{amssymb}  % assumes amsmath package installed
\usepackage{booktabs}
\usepackage{algorithm}
\usepackage[noend]{algpseudocode}
\usepackage{hyperref}

\definecolor{directcolor}{RGB}{17,85,204}    % MATLAB blue
\definecolor{chainedcolor}{RGB}{255,0,255}  % MATLAB purple

\definecolor{t_0}{RGB}{255,50,50}
\definecolor{t_1}{RGB}{210,180,0} 
\definecolor{t_current}{RGB}{1,191,220}

\definecolor{new_chained_connction}{RGB}{40,200,40}

\definecolor{fastlio}{RGB}{228,26,28}
\definecolor{lamm}{RGB}{55,126,184}
\definecolor{unchained}{RGB}{77,175,74}
\definecolor{chained}{RGB}{152,78,163}
\definecolor{gt}{RGB}{255,127,0}

\title{\LARGE \bf
Chain-SLAM: Globally Consistent Backend for Multi-Session LiDAR SLAM via Chained Loop Closure
}

\author{
Zhiheng Li$^{1}$, 
Xinhao Liu$^{1}$, 
Juexiao Zhang$^{1}$, 
Yongqing Liang$^{1\dagger}$, 
Chen Feng$^{1\dagger}$%
\thanks{$^{1}$New York University.}
\thanks{$^{\dagger}$Equal advising.}
}

\newcommand{\modelname}[1]{{Chain-SLAM}}

\begin{document}

\maketitle
\thispagestyle{empty}
\pagestyle{empty}

%%%%%%%%%%%%%%%%%%%%%%%%%%%%%%%%%%%%%%%%%%%%%%%%%%%%%%%%%%%%%%%%%%%%%%%%%%%%%%%%
\begin{abstract}

% Maintaining geometric consistency over long spatial and temporal horizons remains a key challenge in large-scale LiDAR SLAM, particularly when integrating maps collected across multiple sessions. In this work, we present \textbf{\modelname}, a LiDAR SLAM backend that enables globally consistent online map reuse and alignment across multiple sessions with repeated traversals.
% Maintaining geometric consistency over long spatial and temporal horizons remains a fundamental challenge in large-scale LiDAR SLAM, especially when integrating maps collected across multiple sessions. We present \textbf{\modelname}, a LiDAR SLAM backend enabling online multi-session map reuse and alignment with global consistency at large scale.
% We implement a chained loop closure mechanism that efficiently propagates geometric constraints across connected keyframes, enabling robust long-horizon consistency triggered by reliable short-horizon loop closures. The system supports on-the-fly alignment of loaded maps and newly acquired trajectories within a unified pose graph, maintaining both inter- and intra-session geometric consistency without requiring explicit dynamic object removal. Experimental results demonstrate improved trajectory accuracy and robust multi-session integration on large-scale datasets. We release our dataset, code, and framework to support reproducible research in multi-session SLAM and large-scale 3D reconstruction.

Maintaining consistency over long spatial and temporal horizons remains a fundamental challenge in large-scale LiDAR SLAM, particularly when integrating maps collected across multiple sessions. We present \textbf{\modelname}, a LiDAR SLAM backend enabling online multi-session map alignment and reuse with global consistency at large scale. We implement a chained loop closure mechanism that efficiently propagates geometric constraints across inter-session keyframes through an adjacency graph, enabling robust long-horizon consistency triggered by reliable short-horizon loop closures. The system initializes inter-session alignment with GNSS-proximity place recognition, then performs on-the-fly loop closure detections and joint optimization of loaded maps and newly acquired trajectories within a unified factor graph, maintaining both inter- and intra-session geometric consistency without dynamic object removal, and cross-platform robustness with minimal hyperparameter tuning. Experimental results show improved trajectory accuracy and robust multi-session integration on large-scale datasets. We release our source code to support reproducible research in large-scale multi-session LiDAR SLAM. Project site: \href{https://ai4ce.github.io/Chain-SLAM/}{https://ai4ce.github.io/Chain-SLAM/}

\end{abstract}

%%%%%%%%%%%%%%%%%%%%%%%%%%%%%%%%%%%%%%%%%%%%%%%%%%%%%%%%%%%%%%%%%%%%%%%%%%%%%%%%
\section{INTRODUCTION}

Large-scale SLAM is fundamental for robotic systems supporting various applications such as autonomous driving, infrastructure inspection, and field robotics. While recent advances achieve high-fidelity mapping in small environments, robust long-term solutions for large-scale settings remain limited. Large environments are often mapped across multiple sessions, and long spatio-temporal horizons introduce significant variations during revisits due to dynamic objects, lighting, or weather changes. A map built in earlier sessions can therefore provide valuable priors for new sessions, improving robustness against such environmental changes. However, to the best of our knowledge, no open-source SLAM system currently supports real-time multi-session mapping or on-the-fly reuse of previously built maps.

Existing SLAM systems achieve strong local accuracy through tightly coupled LiDAR– or visual–inertial odometry with loop closures and factor graphs for global optimization, but lack support for multi-session deployment or map reuse. Conversely, existing map merging methods treat merging as a post-processing step on individually built maps. As a result, they fail to exploit prior session information during operation and struggle to scale to large scenes with many revisits, where global optimization must simultaneously satisfy a large number of constraints. Furthermore, the performance of existing SLAM and map merging systems varies across hardware platforms, often relying on high-fidelity sensors, high-rate measurements, or extensive hyperparameter tuning.

These limitations motivate an incremental approach to map merging and global optimization for robust large-scale multi-session SLAM. In particular, a real-time map-reuse backend is needed to manage numerous inter- and intra-session loop closures while balancing global alignment and local odometry consistency. Since long spatio-temporal horizons also introduce significant environmental changes that challenge place recognition, robust and efficient long-term correspondence discovery is equally critical. Finally, cross-platform robustness is essential to ensure generalization across diverse robotic platforms and deployment conditions.

\begin{figure}[t]
\centering
\includegraphics[width=\columnwidth]{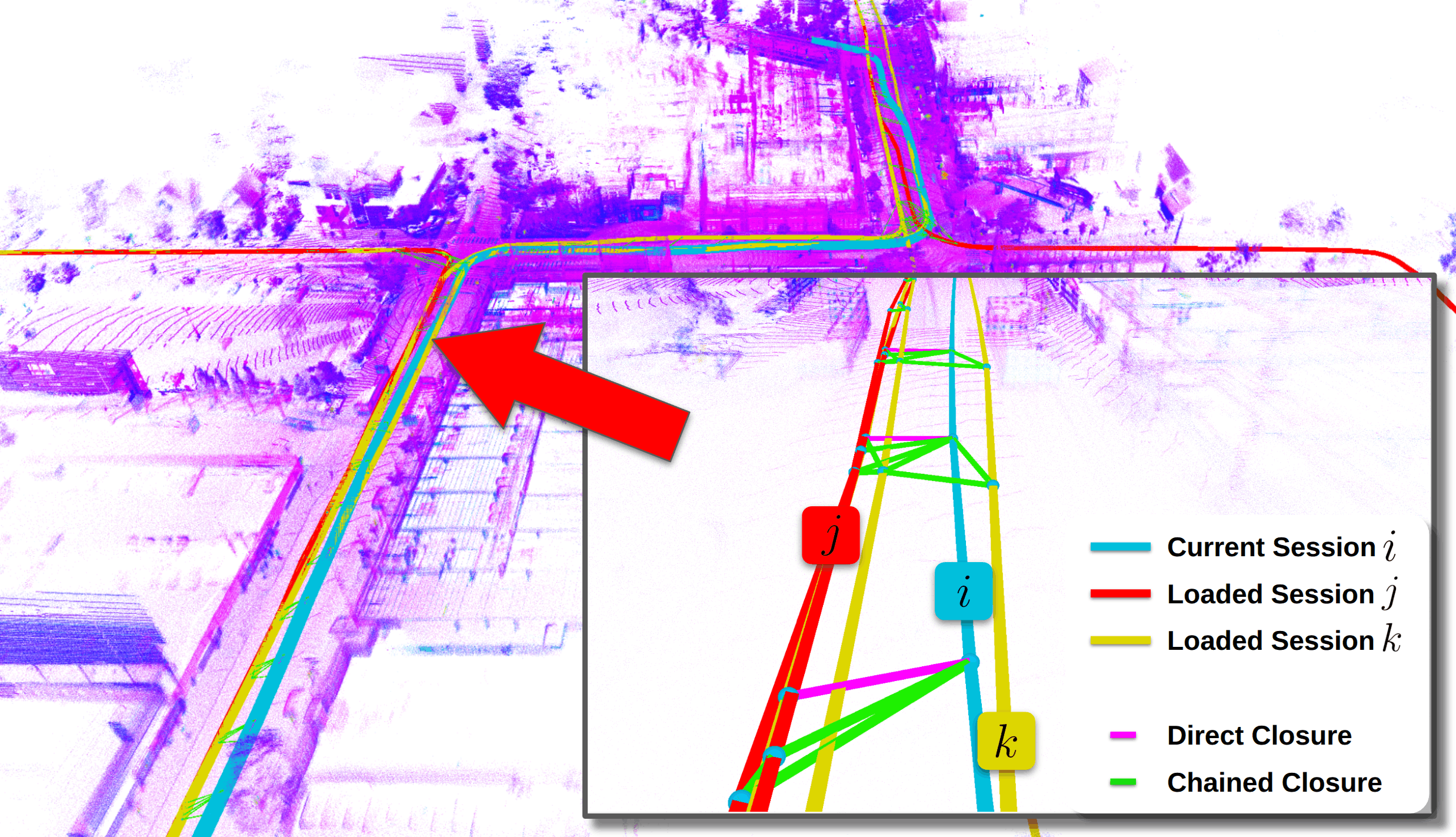}
\caption{
\textcolor{chainedcolor}{\textbf{Direct}} and 
\textcolor{new_chained_connction}{\textbf{Chained}} loop closures between \textcolor{t_current}{\textbf{current session $i$}}, \textcolor{t_0}{\textbf{loaded session $j$}} and \textcolor{t_1}{\textbf{loaded session $k$}}.
Each successful closure stemming from current session triggers the establishment of multiple chained closures between current and loaded sessions, enhancing multi-session consistency.
% between (\textcolor{t_current}{$i$}, \textcolor{t_0}{$j$}), and (\textcolor{t_current}{$i$}, \textcolor{t_1}{$k$}).
}
\label{fig:chained_closure_demo}
\vspace{-4mm}
\end{figure}

To address these challenges, we present \textbf{\modelname}, a scalable real-time multi-session LiDAR–inertial SLAM system that jointly optimizes trajectories across long spatio-temporal horizons. Built on a tightly coupled LiDAR–inertial odometry frontend and a global factor graph backend, the system supports \textbf{(1)} loading existing maps, \textbf{(2)} on-the-fly GNSS-based map merging and alignment, and \textbf{(3)} incremental map reuse and global optimization. 
We further introduce \textbf{chained loop closure} (Fig.~\ref{fig:chained_closure_demo}), a simple yet effective approach to \textbf{(4)} enhance long-term inter-session correspondence discovery by propagating closure constraints through a connectivity graph (Fig.~\ref{fig:adjacency_graph}). \textbf{(5)} Cross-platform robustness is achieved with minimal hyperparameter tuning through scan-to-submap registration for loop closing and submap-to-submap registration for map merging. 
% Finally, we will release our source code to support the research community.
We have released the source code on our project site.

We evaluate the system on two large-scale multi-session datasets, MARS \cite{li2024multiagent} and NCLT \cite{carlevaris2016university}, covering over 20 hours of operations across diverse sensing and motion conditions. 
% Experimental results show improvements in trajectory accuracy, geometric consistency, and viewpoint alignment compared to existing methods with minimal hyperparameter tuning across platforms, while maintaining real-time performance.
Experimental results show improved trajectory accuracy, geometric consistency, and viewpoint alignment over existing methods, with minimal cross-platform hyperparameter tuning and real-time performance.

\section{RELATED WORKS}
\label{sec:related_works}

\subsection{Multi-Session SLAM and Map Merging}

% Multi-session (or multi-map) SLAM aims to align trajectories collected at different times into a consistent global frame. In LiDAR-based systems, a common pipeline consists of (i) inter-session place recognition, (ii) rigid alignment through geometric registration, and (iii) global consistency enforcement via pose graph optimization. For example, LT-Mapper detects inter-session correspondences using Scan Context and aligns sessions using ICP and pose graph optimization \cite{kim2022lt}. Similarly, LAMM establishes inter-map correspondences using global descriptors and estimates transformations through geometric registration to merge independently built maps \cite{wei2024large}. However, these approaches operate primarily as post-processing steps after individual sessions are fully constructed. This limits immediate reuse of previously built maps and makes global optimization increasingly difficult to converge at large scale, where the number of inter-session correspondences can grow substantially.
Multi-session (or multi-map) SLAM aims to align trajectories collected at different times into a consistent global frame. In LiDAR-based systems, this typically involves inter-session place recognition, geometric registration, and pose graph optimization for global consistency. For example, LT-Mapper detects inter-session correspondences using Scan Context and aligns sessions via ICP and pose graph optimization \cite{kim2022lt}, while LAMM uses global descriptors and geometric registration to merge independently built maps \cite{wei2024large}. However, these approaches operate primarily as post-processing on finished individual sessions, limiting immediate reuse of prior maps and making global optimization harder to converge at large scale due to increasing number of inter-session constraints.

Related multi-agent SLAM systems maintain multi-map consistency through inter-agent place recognition, relative pose estimation, and joint pose graph optimization, enabling cross-agent relocalization and map alignment \cite{tian2022kimera, chang2022lamp}. However, they typically assume limited temporal gaps with minor environmental changes. In long-term multi-session settings, appearance variation, structural changes, and dynamic objects degrade place recognition and registration. Some methods mitigate this via dynamic object removal or motion segmentation \cite{lim2021erasor, kim2020remove, mersch2022receding}, but these additions increase system complexity and limit real-time performance.

% In the visual and visual-inertial domain, multi-map representations developed for handling tracking failures provide useful insights for long-horizon multi-map consistency. ORB-ATLAS maintains multiple submaps and merges them when place recognition establishes cross-map correspondences, enabling online map reuse and consistent global optimization \cite{elvira2019orbslam}. ORB-SLAM3 harnesses this framework by creating a new submap when tracking is lost and subsequently merging it back into an existing map once relocalization is achieved, followed by joint optimization over the unified graph \cite{campos2021orb}.
In the visual-inertial domain, multi-map representations developed for handling tracking failures offer insights for long-horizon consistency. ORB-ATLAS maintains multiple submaps and merges them upon detecting cross-map correspondences, enabling online map reuse and joint optimization \cite{elvira2019orbslam}. ORB-SLAM3 extends this framework by creating new submap when tracking is lost and merging it back after relocalization, followed by global optimization over the unified graph \cite{campos2021orb}.

% Inspired by these approaches, our system enables online multi-session map integration by treating a loaded prior map as an existing base map and representing the incoming session as a new submap. Upon detecting inter-session correspondences, the new session is relocalized and aligned to the base map, after which both maps are merged into a unified pose graph and jointly optimized, thus achieving globally consistent multi-session SLAM.
% The chained loop closures efficiently establish inter-session registration constraints without the need for exhaustive individual place recognitions, avoiding potential match failures due to environment changes while eliminating the need for explicit dynamic object removal. 

Inspired by these approaches, our system enables online multi-session map integration by treating a loaded prior map as an existing base map and representing the incoming session as a new submap. Upon detecting inter-session correspondences, the new session is relocalized and aligned to the base map, after which both maps are merged into a unified pose graph and jointly optimized, achieving globally consistent multi-session SLAM. 
Our chained loop closure (Sec.~\ref{subsec:chained_loop_closure}) establishes additional inter-session constraints by propagating correspondences through connected keyframes, reducing correspondence search cost and avoiding false negatives from place recognition failures.
This mechanism improves robustness to environmental changes while eliminating the need for explicit dynamic object removal and maintaining real-time performance.

\subsection{Sequential and Multi-Constraint Data Association}
\label{subsec:rw_multiassoc}

A second line of work improves robustness through \emph{sequence-level} or \emph{multi-constraint} association, inspiring our chained loop closure design for enhanced global consistency. In visual place recognition (VPR), sequential methods such as JIST aggregate evidence over time rather than relying on a single query frame, improving robustness to perceptual aliasing and appearance changes \cite{berton2023jist}.
% JIST replaces one-to-one frame matching with sequence-level association by aggregating temporally connected observations into compact descriptors
Recent attention-based methods further model spatio-temporal persistence for more discriminative place representations under challenging viewpoint and environmental changes \cite{kiu2025flexible, mereu2022learning}. Beyond temporal aggregation, panoramic inputs provide richer context by exploiting panoramic structure and self-supervision to refine pseudo-labels and improve viewpoint robustness \cite{chen2024self}.

In SLAM back-ends, robustness to noisy or incorrect associations can be improved by enforcing \emph{multi-edge geometric consistency} rather than trusting individual correspondences. Graph-theoretic and consistency-maximization methods construct compatibility graphs and select consistent subsets of loop closures or measurements, using global compatibility to identify inliers and suppress outliers. These approaches leverage dense subgraph selection or maximum clique formulations over geometrically invariant measurements to prune inconsistent correspondences and enable robust alignment under extreme outlier rates \cite{lusk2021clipper, yang2020teaser}. 

% While prior consistency-maximization methods focus on rejecting incorrect associations among many candidates, sequential VPR approaches show that robustness can also be improved by aggregating temporally or spatially related constraints. Inspired by these complementary perspectives, we expand geometric validation from reliable seed correspondences. Each inter-session loop closure triggers rapid discovery of connected keyframes through a unified adjacency graph, enabling additional geometrically consistent constraints without exhaustive candidate search.
% These augmented constraints effectively enforces bidirectional cross-validation between the new session and the loaded map: previously mapped structures stabilize the new session against local estimation errors, while newly acquired observations introduce corrective constraints that reduce accumulated drift in the existing map. 
% Therefore, our chained loop closure improves global consistency and reduces absolute trajectory error with minimal computational overhead.

While prior consistency-maximization methods focus on rejecting incorrect associations among many candidates, these methods show robustness can also be improved by aggregating temporally or spatially related constraints. Inspired by such perspectives, we expand geometric validation from reliable seed correspondences in SLAM backend: each inter-session loop closure triggers discovery of connected keyframes through a unified adjacency graph, enabling additional inter-session constraints without exhaustive candidate search. These constraints provide bidirectional cross-validation between the new session and the loaded map: existing structures stabilize the new session against local estimation errors, while new observations introduce corrections that reduce drift in the existing map. As a result, chained loop closure improves global consistency and reduces trajectory error with minimal overhead.

% \begin{figure}[t]
% \centering
% \includegraphics[width=\columnwidth]{fig/pipeline_v6.png}
% \caption{Frontend--backend multi-session LiDAR--inertial SLAM framework. 
% $\mathbf{x}$ denotes keyframes from current session $i$ and loaded session $j$.
% % $\mathbf{x}_i$ denotes keyframes from the current session, and $\mathbf{x}_j$ denotes keyframes from loaded session(s). 
% The frontend predicts the initial state $\mathbf{x}_i^0$ via IMU propagation and refines it to the odometry estimate $\mathbf{x}_i'$ via LiDAR--inertial EKF. The backend performs incremental factor graph optimization to produce the final state $\mathbf{x}_i^{*}$, jointly updating loaded map states $\mathbf{x}_j^{*}$ to $\mathbf{x}_j^{**}$ through \textcolor{directcolor}{direct} and \textcolor{chainedcolor}{chained} loop closures.}
% \label{fig:pipeline}
% \end{figure}

\begin{figure*}[t]
\centering
\includegraphics[width=0.95\textwidth]{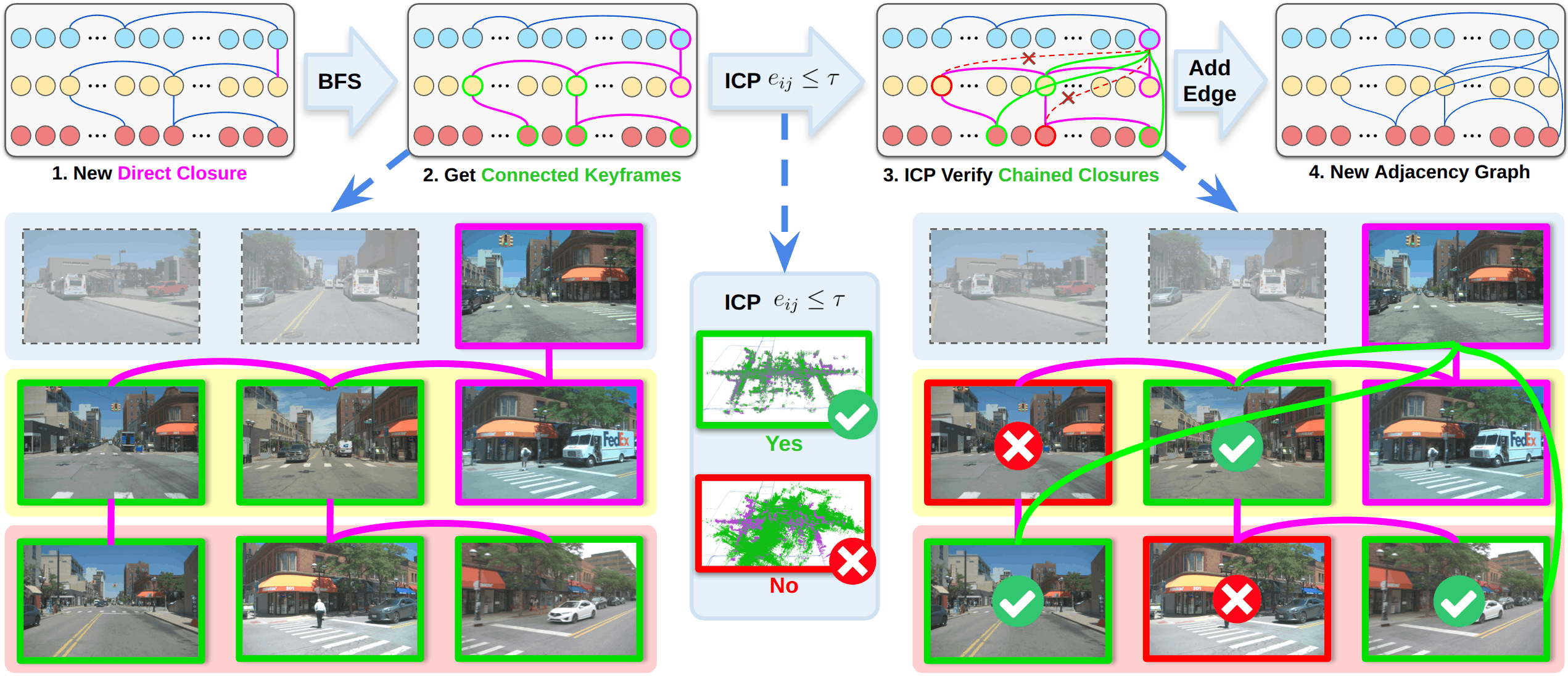}
\caption{Chained loop closure detection via a cross-session adjacency graph. Nodes are keyframes, each row corresponds to \textcolor{t_0}{\textbf{session 0}}, \textcolor{t_1}{\textbf{session 1}}, and \textcolor{t_current}{\textbf{current session}}. 
Edges denote loop closures: \textcolor{directcolor}{\textbf{finalized}}, \textcolor{chainedcolor}{\textbf{traced existing connections}}, and \textcolor{unchained}{\textbf{accepted chained closures}}. 
\textbf{Step 1:} A direct inter-session closure is detected. 
\textbf{Step 2:} BFS retrieves connected keyframes as chained candidates. 
\textbf{Step 3:} A chained closure between $i$ and $j$ is accepted if scan-to-submap ICP error $e_{ij} \le \tau$ . Temporal adjacency is filtered by regular loop closure criteria (Sec.~\ref{subsec:loop_closure}).
\textbf{Step 4:} Accepted closures are updated to the graph.
}
\label{fig:adjacency_graph}
\vspace{-6mm}
\end{figure*}
\section{Method}

% \subsection{System Overview}

% We develop a LiDAR--inertial SLAM system that supports multi-session and multi-map operation within a unified factor graph framework. The system follows a frontend--backend architecture. The frontend performs tightly coupled LiDAR--inertial odometry using a formulation based on FAST-LIO2, which estimates motion through direct scan-to-map registration. The back end is based on Fast\_LIO\_SAM which integrates LIO\_SAM style factor graph based optimizations with Fast-LIO2 frontend. An incremental k-d tree structure maintains the local map and enables efficient nearest-neighbor queries for point-to-map correspondence search.

% The backend maintains a global factor graph that jointly optimizes keyframe poses across sessions. Each keyframe stores a six-degree-of-freedom (6-DoF) pose and an associated submap. Odometry factors connect consecutive keyframes, while loop closure and map alignment factors introduce geometric constraints between non-consecutive keyframes. Previously constructed maps can be loaded and appended to the graph, allowing joint optimization of trajectories collected at different times while maintaining consistent global geometry.
\subsection{System Overview}

We develop a LiDAR--inertial SLAM system that supports multi-session and multi-map operation within a unified factor graph framework. 
% Figure~\ref{fig:pipeline} shows an overview of the proposed system.
This system follows a frontend--backend architecture. The frontend performs tightly coupled LiDAR--inertial odometry using a formulation derived from FAST-LIO2~\cite{xu2022fast}, which estimates motion through direct scan-to-map registration against an incrementally maintained local map by a K-D Tree.
% An incremental K-D Tree structure enables efficient nearest-neighbor queries for point-to-map correspondence, supporting accurate and real-time state estimation.
The backend extends the Fast\_LIO\_SAM \cite{wang2022fast_lio_sam} framework by incorporating chained loop closure and map merging within a single LIO\_SAM-style factor graph optimization framework to enable multi-session operation and improve global geometric consistency \cite{shan2020lio}. 
In our system, previously constructed maps can be loaded and appended to the factor graph, allowing all sessions to be jointly optimized within a unified probabilistic framework. 
This approach ensures consistent global alignment across sessions and enables scalable long-term mapping.

We formulate SLAM as a factor graph defined over $N$ keyframe poses, 
$\mathbf{X} = \{ \mathbf{x}_0, \mathbf{x}_1, \dots, \mathbf{x}_N \}, \quad \mathbf{x}_i \in SE(3)$.
Our graph contains Unary Prior Factors and Binary Between Factors.
\textbf{Unary Prior factors} $\mathbf{T}_i$ anchor the starting pose of each trajectory segment:
$
\phi_p(\mathbf{x}_i)
=
\left\|
\log \left(
\mathbf{T}_i^{-1} \mathbf{x}_i
\right)
\right\|_{\mathbf{\Sigma}_p}^2.
$
\textbf{Binary Between factors} $\mathbf{T}_{ij}$ encode relative pose constraints between keyframes. These include both odometry factors and loop closure factors, and share the same mathematical form:
$
\phi_b(\mathbf{x}_i, \mathbf{x}_j)
=
\left\|
\log \left(
\mathbf{T}_{ij}^{-1}
\mathbf{x}_i^{-1}
\mathbf{x}_j
\right)
\right\|_{\mathbf{\Sigma}_{ij}}^2.
$

The Frontend LiDAR--inertial odometry estimates relative motion as odometry factors to connect consecutive keyframes.
The loop closure factors connect non-consecutive keyframes using geometric registration.
The optimal trajectory is obtained by minimizing the objective function:
\vspace{-1mm}
\begin{equation}
\mathbf{X}^*
=
\arg\min_{\mathbf{X}}
\sum \phi_p
+
\sum \phi_b.
\end{equation}
\vspace{-0.5mm}
Incremental optimization is performed using iSAM2 \cite{kaess2012isam2}, which efficiently updates only affected portions of the graph when new factors are added. Pose corrections are propagated throughout the graph, ensuring global geometric consistency.

\subsection{Loop Closure Detection and Registration}
% \label{subsec:loop_closure}
% Loop closure candidates are identified through spatial proximity search within a predefined radius over keyframe positions indexed in a K-D Tree, with the precondition of candidates being more than a time interval away from each other. Given a candidate keyframe pair $(i, j)$, geometric registration is performed using Iterative Closest Point (ICP) between the single scan of current keyfame $i$ and the submap of matched past keyframe $j$. ICP estimates the rigid transformation $T$ by minimizing the ICP mean squared error $MSE_{ij}$,
% % \vspace{-20pt}
% \vspace{-1.6mm}
% \begin{align}
% \mathbf{T}_{ij} &= 
% \arg\min_{\mathbf{T}}
% MSE_{ij} (\mathbf{T}),\\
% MSE_{ij} (\mathbf{T}) &= 
% \frac{1}{|\mathcal{C}|}
% \sum_{(\mathbf{p}, \mathbf{q}) \in \mathcal{C}}
% \left\|
% \mathbf{p} - \mathbf{T}\mathbf{q}
% \right\|^2,
% \end{align}
% \vspace{-2mm}

% % \vspace{-5pt}
% where $\mathcal{C}$ denotes the set of inlier point correspondences of the keyframe pair $(i,j)$. 
% A loop closure is accepted only if ICP converges and an error score $f_{ij}=\frac{\#inlier}{\#target\ points}$ is greater than a threshold $f_{ij} \ge \tau$.
% Accepted loop closures introduce a factor with covariance derived from ICP error score $\mathbf{\Sigma}_{ij}=\lambda f_{ij} \mathbf{I}_6$,
% where $\lambda$ is a scaling factor. This assigns higher confidence to geometrically consistent alignments. Loop closure factors reduce drift and enforce geometric consistency across revisited regions.

\label{subsec:loop_closure}
Loop closure candidates are identified through spatial proximity search within a predefined radius over keyframe positions indexed in a K-D Tree, with the additional constraint that candidate frames are separated by a minimum temporal interval. Given a candidate keyframe pair $(i, j)$, geometric registration is performed using Iterative Closest Point (ICP) between the single scan of the current keyframe $i$ and the submap of the matched past keyframe $j$. 
ICP estimates the rigid transformation $\mathbf{T}_{ij}$ by minimizing the mean squared alignment error
\vspace{-1.6mm}
\begin{align}
\mathbf{T}_{ij} &= 
\arg\min_{\mathbf{T}} MSE(\mathbf{T}),\\
MSE(\mathbf{T}) &= 
\frac{1}{|\mathcal{C}^{\mathrm{I}}_{ij}|}
\sum_{(\mathbf{p}, \mathbf{q}) \in \mathcal{C}^{\mathrm{I}}_{ij}}
\left\|
\mathbf{p} - \mathbf{T}\mathbf{q}
\right\|^2,
\end{align}
where $\mathcal{C}^{\mathrm{I}}_{ij}$ denotes the set of ICP inlier correspondences for the keyframe pair $(i,j)$ and $\mathcal{P}_t$ denotes the set of target points in the submap. Let $e_{ij}=MSE(\mathbf{T}_{ij})$ denote the minimized error. A loop closure is accepted only if ICP converges and $e_{ij} \le \tau$, where $\tau$ is a predefined error threshold. Accepted loop closures introduce a factor with covariance derived from the ICP error, $\mathbf{\Sigma}_{ij}=\lambda e_{ij}\mathbf{I}_6$, where $\lambda$ is a scaling factor. This assigns higher confidence to geometrically consistent alignments. Loop closure factors reduce drift and enforce geometric consistency across revisited regions.

\subsection{Chained Loop Closure}
\label{subsec:chained_loop_closure}

While direct loop closures correct drift between individually matched keyframes, they lack explicit global optimization of the established loops. As a result, residual inconsistencies may remain across large-scale multi-session maps. In the case of multi-session deployment, the connectivity structure of previously established loop closures can provide rich information for correspondence identification. To this end, we introduce \textbf{Chained Loop Closure}, a graph-based mechanism that propagates geometric constraints beyond a single direct correspondence to enhance global consistency while maintaining local accuracy.

All accepted inter- and intra-session loop closures are maintained in a unified adjacency graph (Fig.~\ref{fig:adjacency_graph}), where each loop closure introduces an edge between two keyframe nodes. A new session without a loaded map initializes the graph using only its intra-session loop closures. When subsequent sessions are loaded, inter-session loop closures detected through the procedure in Sec.~\ref{subsec:loop_closure} introduce edges between current-session keyframes and loaded-map keyframes, thereby joining previously separate connections into inter-session connected components. As more sessions are introduced, inter-session edges incrementally expand the connected components, forming clusters of geometrically proximate keyframes spanning multiple sessions.

This evolving graph structure enables efficient cross-session correspondence retrieval through traversal of connected components rather than repeated global place recognition or spatial search. At the same time, chained loop closure ensures that each direct inter-session match establishes additional constraints within the corresponding connected component, further establishing inter-session edges and strengthening inter-session connectivity. Consequently, each connected component becomes an anchoring unit in the global map, consisting of clusters of intra- and inter-session loop-closed keyframes.
Importantly, chained loop closures are applied only between current session and loaded-map keyframes, not within current session. This design enhances inter-session consistency while avoiding redundant constraints on locally consistent intra-session odometry, maintaining a balance between global and local optimization.

A pseudocode description of the complete chained loop closure process is shown in Algorithm~\ref{alg:chained_loop}.
Specifically, when a direct loop closure is established between the current keyframe $i$ and loaded-map keyframe $j$, we perform a breadth-first search (BFS) on the adjacency graph to identify all loaded keyframes belonging to the same connected component as $j$, producing the connected set
\begin{equation}
\mathcal{N}(j) = \{\, k \mid k \text{ is loop-closure connected to } j \,\}.
\label{eq:connected_set}
\end{equation}

After identifying all connected keyframes through graph traversal $k \in \mathcal{N}(j)$ \eqref{eq:connected_set}, we perform ICP registration between the current keyframe point cloud scan and the submap of $k$. If ICP converges and satisfies the error criterion in Sec.~\ref{subsec:loop_closure}, a chained loop closure factor is added between keyframes $i$ and $k$, with covariance derived from the ICP error score.

\begin{algorithm}[t]

\caption{Chained Loop Closure}
\label{alg:chained_loop}

\begin{algorithmic}[1]
\Require Current keyframe $i$, matched keyframe $j$
\Require Adjacency graph $\mathcal{G}$, loaded keyframe count $N_L$

\If{$j \ge N_L$}
    \Return $j$ is not from loaded map
\EndIf

% \Statex \textbf{/* Step 1: BFS Retrieve connected keyframes */}
\State $\mathcal{N}(j) \gets \{ j \}$ \Comment{\textbf{Step 1: BFS }}

\ForAll{keyframes $k$ reachable from $j$ in $\mathcal{G}$}
    \If{$k < N_L$} \Comment{Add loaded keyframes only}
        \State Add $k$ to $\mathcal{N}(j)$ 
    \EndIf
\EndFor

% \Statex \textbf{/* Step 2: Chained ICP validation \Comment{Chained ICP validation} */}
\ForAll{$k \in \mathcal{N}(j)$} \Comment{\textbf{Step 2: ICP validation}}
    \If{$k = j$}
        \State \textbf{continue} \Comment{Avoid self-matching frame $j$}
    \EndIf
    \State Construct submap of keyframe $k$
    \State Compute $(\mathbf{T}_{ik}, f_{ik})$ via ICP
    \If{$f_{ik} \le \tau$}
        \State $\mathbf{\Sigma}_{ik} \gets \lambda f_{ik}\mathbf{I}_6$ \Comment{Factor covariance from ICP}
        \State Add loop closure factor between $i$ and $k$
    \EndIf
\EndFor

\end{algorithmic}
\end{algorithm}
% \vspace{-4mm}

% \begin{algorithm}[t]
% \centering
% \scalebox{0.96}{
% \begin{minipage}{\columnwidth}
% \caption{Chained Loop Closure}
% \label{alg:chained_loop}
% \begin{algorithmic}[1]
% \Require Current keyframe $i$, matched keyframe $j$
% \Require Adjacency graph $\mathcal{G}$, loaded keyframe count $N_L$

% \If{$j \ge N_L$}
%     \Return $j$ is not from loaded map
% \EndIf

% \Statex \textbf{/* Step 1: BFS Retrieve connected keyframes */}
% \State $\mathcal{N}(j) \gets \{ j \}$

% \ForAll{keyframes $k$ reachable from $j$ in $\mathcal{G}$}
%     \If{$k < N_L$} \Comment{Add loaded keyframes only}
%         \State Add $k$ to $\mathcal{N}(j)$ 
%     \EndIf
% \EndFor

% \Statex \textbf{/* Step 2: Chained ICP validation */}
% \ForAll{$k \in \mathcal{N}(j)$}
%     \If{$k = j$}
%         \State \textbf{continue} \Comment{Avoid self-matching frame $j$}
%     \EndIf
%     \State Construct submap of keyframe $k$
%     \State Compute $(\mathbf{T}_{ik}, f_{ik})$ via ICP
%     \If{$f_{ik} \le \tau$}
%         \State $\mathbf{\Sigma}_{ik} \gets \lambda f_{ik}\mathbf{I}_6$ \Comment{Factor covariance from ICP}
%         \State Add loop closure factor between $i$ and $k$
%     \EndIf
% \EndFor
% \end{algorithmic}
% \end{minipage}
% }
% \end{algorithm}

\subsection{Map Representation and Multi-Session Integration}
\begin{figure}[t]
\centering

\begin{subfigure}{0.9\columnwidth}
    \centering
    \includegraphics[width=\linewidth]{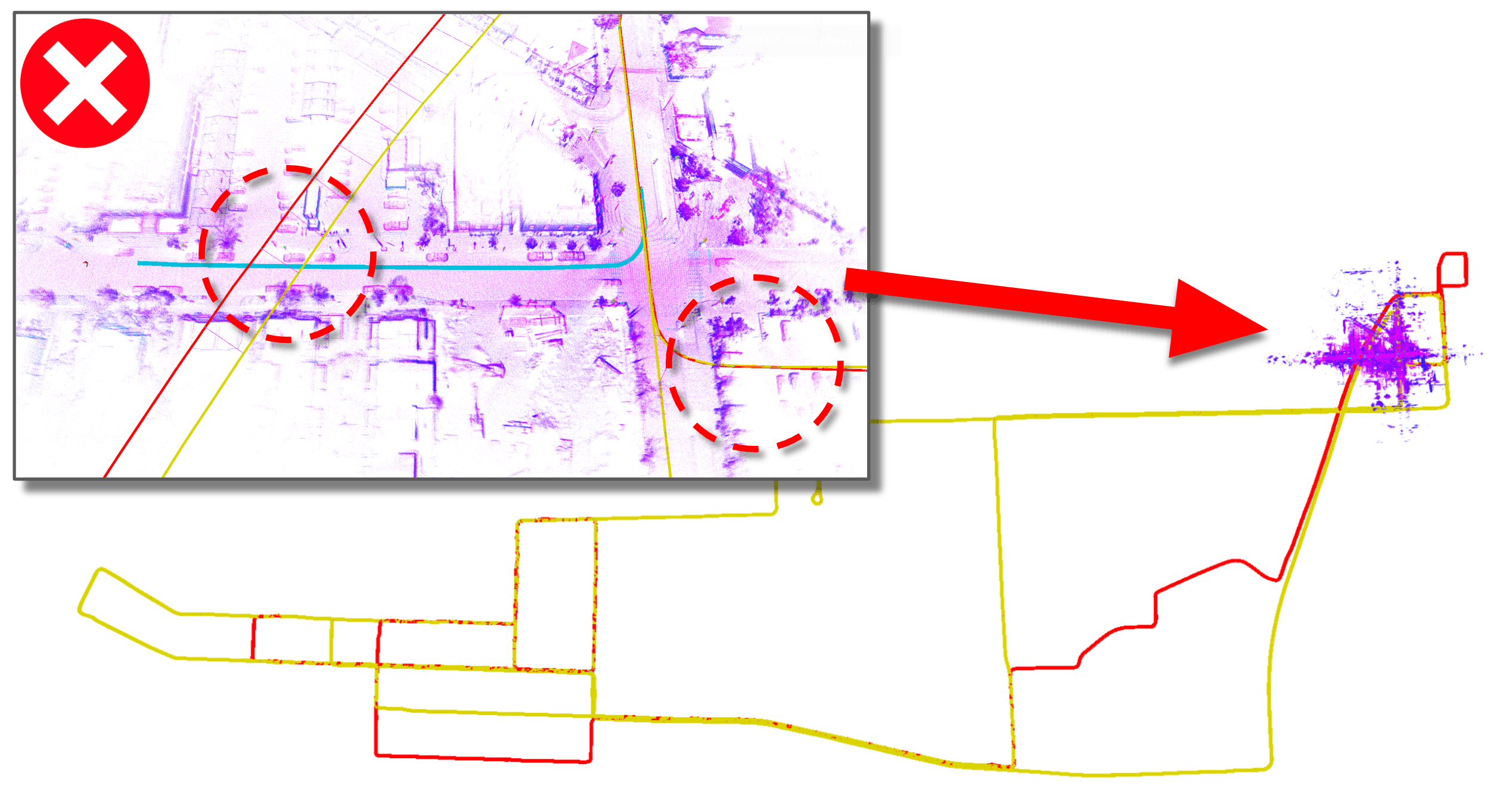}
    \caption{Before map merging, \textcolor{t_current}{\textbf{current session}} initializes trajectory at origin of its own coordinate frame, inconsistent with the map frame established by \textcolor{t_0}{\textbf{session 0}} and \textcolor{t_1}{\textbf{session 1}}.}
    \label{fig:map_align_wrong}
\end{subfigure}

\vspace{0.5em}

\begin{subfigure}{0.9\columnwidth}
    \centering
    \includegraphics[width=\linewidth]{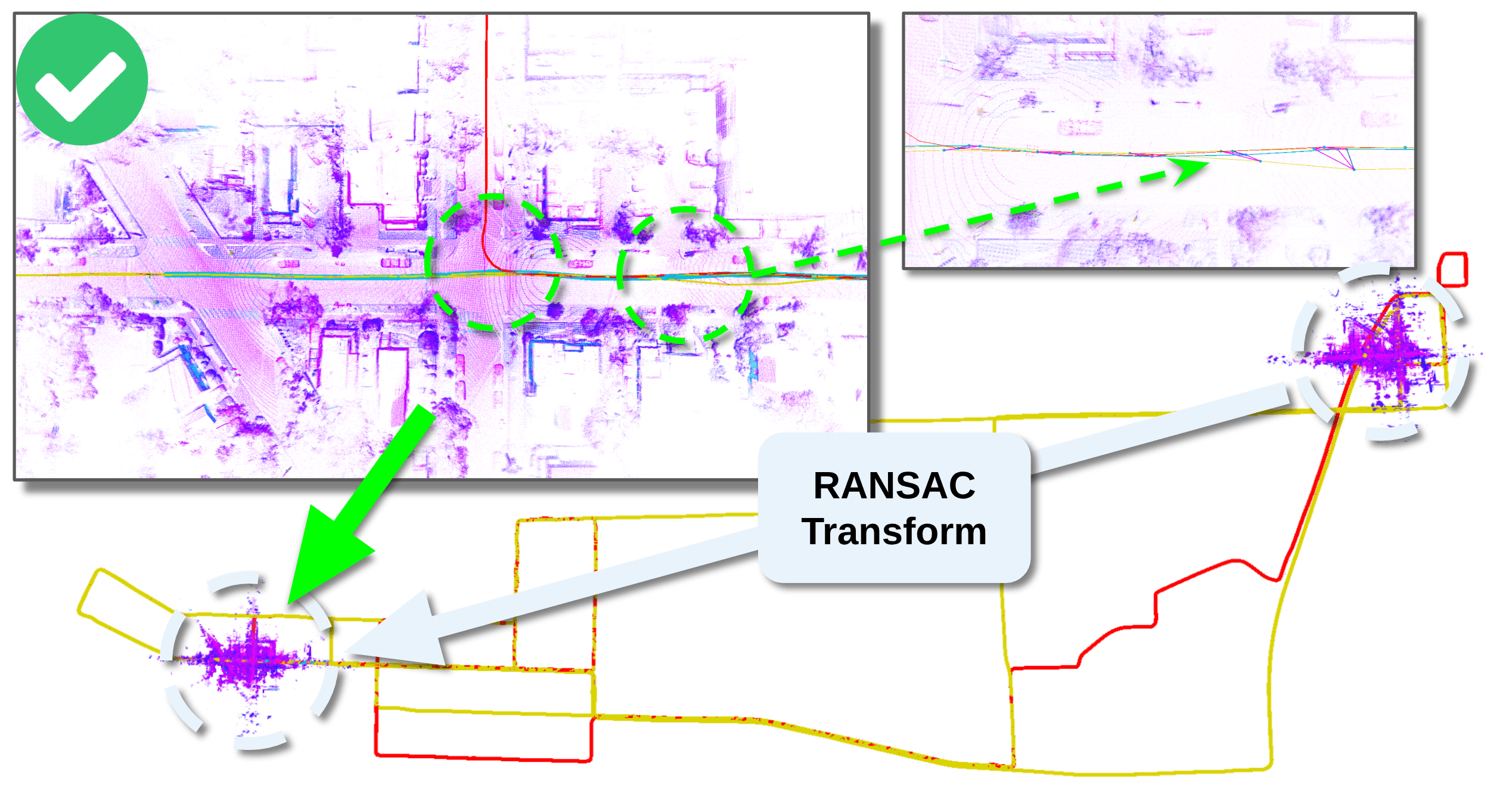}
    \caption{After map merging, RANSAC-estimated transformation aligns \textcolor{t_current}{\textbf{current session trajectory}} to the loaded map frame, enabling re-detection of loop closures with \textcolor{t_0}{\textbf{session 0}} and \textcolor{t_1}{\textbf{session 1}} along the transformed trajectory.}
    \label{fig:map_align_correct}
\end{subfigure}

\caption{Multi-session map alignment}
\label{fig:map_alignment}
\vspace{-6mm}
\end{figure}
Each map is represented as a set of keyframes identified by persistent indices, where each keyframe stores its raw and optimized pose and associated LiDAR scan. 
% Loop closures and scene boundaries are recorded using keyframe index pairs and index ranges, and the closure constraint covariances are stored for each pair. When a map is saved, keyframe poses, LiDAR scans, loop closure relationships, and scene index ranges are stored in this indexed form. 
Scene boundaries are represented as keyframe index ranges, while loop closures are recorded as keyframe index pairs associated with a constraint covariance. When a map is saved, keyframe poses, LiDAR scans, loop closure pairs with their covariances, and scene index ranges are serialized in this indexed representation.
Upon loading, keyframes are restored and appended to the global keyframe container while preserving their indices and session membership.
The corresponding prior, odometry, and loop closure factors are reconstructed in the factor graph with their original covariances. 
This restores the loaded map as a connected component in the graph. New keyframes from the current session are then added incrementally with new indices and connected through odometry factors, forming a separate component initially. Inter-session alignment and loop closures subsequently introduce factors that connect these components, enabling unified optimization and consistent multi-session integration.

Since GNSS is available in most large-scale datasets, we perform place recognition via GNSS proximity search as a stable approach for establishing inter-session correspondence, where keyframe GNSS positions are indexed in a K-D tree for efficient retrieval.
For each candidate pair, feature-based registration using Fast Point Feature Histograms (FPFH) and RANSAC estimates an initial long-horizon transformation $\mathbf{T}_{ij}^{\mathrm{R}}$.
The RANSAC fitness score is defined as the ratio of inlier correspondences to the number of target points $f^{\mathrm{R}}_{ij}=\frac{|\mathcal{C}_{\mathrm{inlier}}|}{|\mathcal{P}_t|}$,
where $\mathcal{C}_{\mathrm{inlier}}$ denotes the set of inlier correspondences and $\mathcal{P}_t$ denotes the target submap points. Higher fitness indicates better alignment, and the transformation is accepted only if
$
f^{\mathrm{R}}_{ij} \ge \tau_{\mathrm{R}},
$
where $\tau_{\mathrm{R}}$ is a predefined threshold. 
% Similar to the ICP acceptance criterion (Sec.~\ref{subsec:loop_closure}), the estimated transformation is accepted only if the RANSAC fitness score satisfies $f^{\mathrm{R}}_{ij} \ge \tau_{\mathrm{R}}$, where $f^{\mathrm{R}}_{ij}=\frac{|\mathcal{C}^{\mathrm{R}}_{ij}|}{|\mathcal{P}_t|}$. Here, $\mathcal{C}^{\mathrm{R}}_{ij}$ denotes the set of RANSAC inlier correspondences, $\mathcal{P}_t$ denotes the set of points in the target submap, and $\tau_{\mathrm{R}}$ is a predefined threshold.
% % The RANSAC fitness score is defined as the ratio of inlier correspondences to the number of target points:
% % \begin{equation}
% % f^{\mathrm{R}}_{ij}
% % =
% % \frac{|\mathcal{C}^{\mathrm{R}}_{ij}|}{|\mathcal{P}_t|},
% % \end{equation}
% % where $\mathcal{C}^{\mathrm{R}}_{ij}$ denotes the set of RANSAC inlier correspondences and $\mathcal{P}_t$ denotes the target submap points. The estimated transformation is accepted only if $f^{\mathrm{R}}_{ij} \ge \tau_{\mathrm{R}}$, where $\tau_{\mathrm{R}}$ is a predefined threshold.

As shown in Fig.~\ref{fig:map_alignment}, the first accepted transformation is applied to align current trajectory with loaded map.
Following alignment, the factor graph is rebuilt to ensure consistency between the transformed trajectory and optimization framework. In particular, the prior factor anchoring the current session is replaced with one consistent with the aligned pose, and the graph is reinitialized using transformed poses and updated constraints. 
Additional loop closures are then detected along the aligned trajectory using proximity search and ICP registration, following the procedures described in Sections~\ref{subsec:loop_closure} and~\ref{subsec:chained_loop_closure}.
These alignment and loop closure factors merge the current session and the loaded map into a unified factor graph, enabling subsequent incremental joint optimization and on-the-fly multi-session map reuse.

\subsection{Hyperparameter Considerations}
While our system contains several tunable hyperparameters, extensive tuning is not required for cross-platform robustness. Despite the substantial platform differences between the MARS and NCLT datasets (Tab.~\ref{tab:dataset_stats}), both use identical hyperparameters except for (1) sensor-specific parameters, including LiDAR channel number, IMU noise characteristics, and LiDAR--IMU extrinsics, and (2) loop closure detection frequency and keyframe generation threshold, which are adjusted according to platform speed. 
The loop closure ICP error threshold $\tau$ and the FPFH and RANSAC settings for map merging are identical for both datasets.
% The loop closure ICP error threshold $\tau$ is fixed at 0.3 for both datasets, and the FPFH and RANSAC settings for GNSS-based map merging remain identical.

Registration robustness primarily comes from submap-based alignment. Submaps are constructed by aggregating keyframes within an index radius $n$ around a center keyframe, resulting in up to $2n+1$ keyframes per submap. When fewer neighboring frames are available, smaller submaps are used and unreliable matches are rejected by the ICP error threshold. Loop closure uses scan-to-submap ICP, aligning the keyframe scan to the submap of a matched historical keyframe (Sec.~\ref{subsec:loop_closure}). For the more noise-sensitive FPFH+RANSAC registration in map merging, we perform submap-to-submap registration: the current submap aggregates up to $n+1$ keyframes (no future frames available) while the historical submap aggregates up to $2n+1$. All point clouds are voxel-downsampled prior to registration. We used the same $n$ and voxel sizes for both datasets.

The loop closure detection frequency is adjusted according to platform speed to maintain spatial separation between loop closure keyframes and prevent excessive growth of connected components in the adjacency graph. 
Although temporally adjacent keyframes are already excluded by a time gap requirement within a single session (Sec.~\ref{subsec:loop_closure}), multi-session integration can create indirect connections between consecutive keyframes through chained closures with other sessions, producing locally dense closure constraints that may dominate odometry factors.

The root cause of this issue is that to ensure chained candidate retrieval across arbitrarily large number of sessions, the BFS imposes no explicit depth limit (Eq.~\ref{eq:connected_set}). 
As a result, its effective propagation range is only implicitly governed by the loop closure detection radius.
If closures are detected too frequently relative to this radius, the spatial coverage of different closures may overlap, merging multiple components in the adjacency graph. In the extreme case, the entire multi-session map can become a single connected component. To avoid this, the loop closure detection frequency $f_{\text{lc}}$ should be chosen with platform speed $v$ and detection radius $r_{\text{lc}}$ such that the robot travels beyond the radius between detections $\frac{v}{f_{\text{lc}}} > r_{\text{lc}}.$
% \[
% \frac{v}{f_{\text{lc}}} > r_{\text{lc}}.
% \]
This maintains spatially separated components and keeps the effective BFS search range comparable to the direct closure detection radius.

\begin{figure*}[t]
\centering
\includegraphics[width=0.8\textwidth]{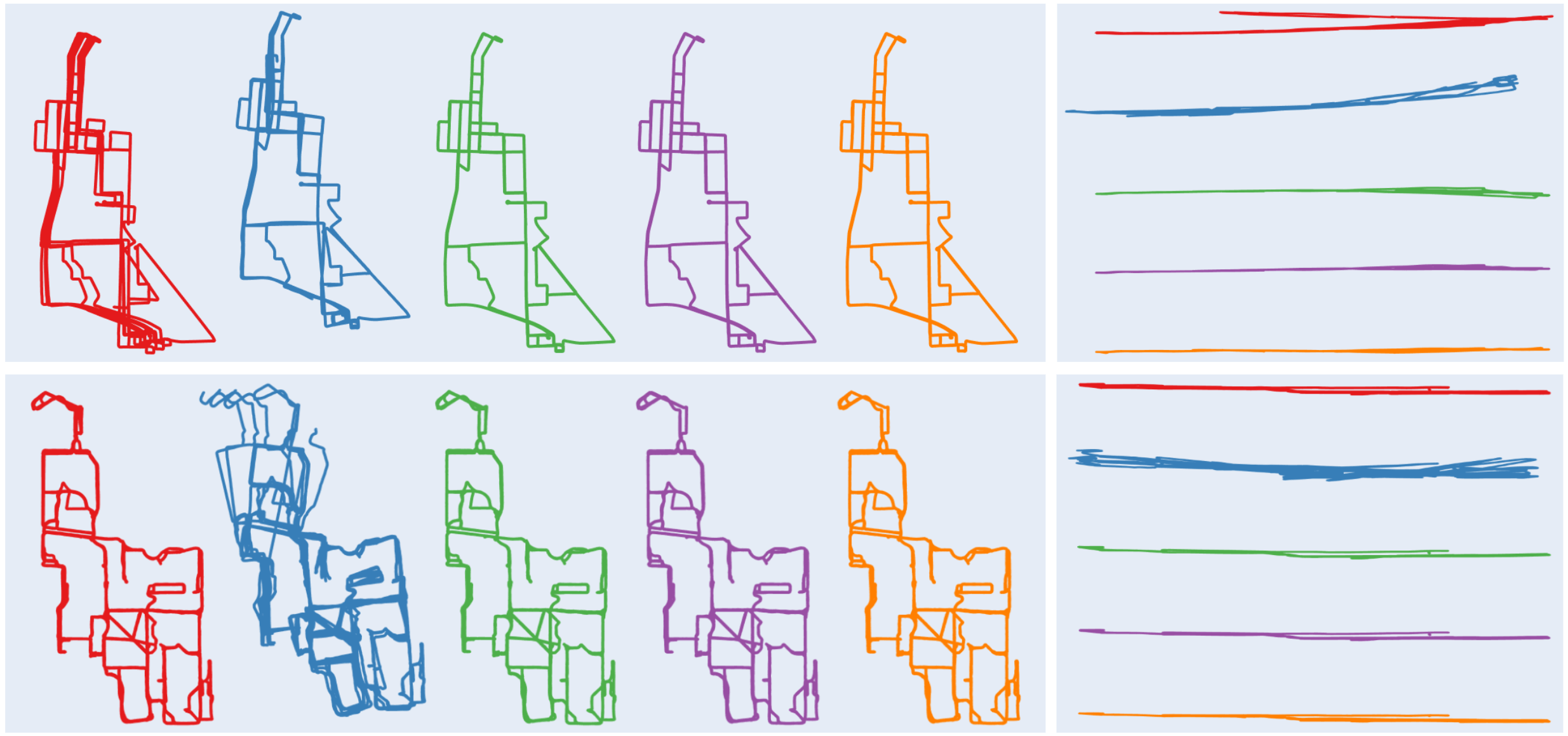}
% \caption{
% Trajectory results no MARS (top) and NCLT (bottom) from different methods:
% \textcolor{fastlio}{\textbf{Fast-LIO2}},
% \textcolor{lamm}{\textbf{LAMM}},
% \textcolor{unchained}{\textbf{Ours Unchained}},
% \textcolor{chained}{\textbf{Ours Chained}}, and
% \textcolor{gt}{\textbf{Ground Truth}}.
% \textbf{Fast-LIO2:} no loop closure implemented, resulting in misalignments at revisits especially on large high-speed scenes.
% \textbf{LAMM:} global optimizer struggles to optimize hundreds of loop closures simultaneously in post-processing, resulting in distortions and misalignments.
% \textbf{Ours unchained:} overall well aligned from top view, but slight distortion exists at long roads with limited revisits.
% \textbf{Ours chained:} best quality with almost no distortion anywhere.
% }
\caption{
Trajectory results on MARS (top) and NCLT (bottom):
\textcolor{fastlio}{\textbf{Fast-LIO2}},
\textcolor{lamm}{\textbf{LAMM}},
\textcolor{unchained}{\textbf{Ours Unchained}},
\textcolor{chained}{\textbf{Ours Chained}}, and
\textcolor{gt}{\textbf{Ground Truth}}.
\textbf{Fast-LIO2:} no loop closure, misalignments at revisits, especially in large high-speed scenes.
\textbf{LAMM:} global optimization over many loop closures introduces distortions and misalignments.
\textbf{Ours unchained:} generally well aligned, with slight distortion along long roads with limited revisits.
\textbf{Ours chained:} best alignment with minimal distortion.
}
\label{fig:trajectory}
\vspace{-3mm}
\end{figure*}

\section{EXPERIMENTS}
\label{sec:experiments}

% \yq{I have a concern that, we didn't explicitly say the settings of the two variants: unchained and chained. May be we can put them in the last subsection (hyper-param) or the beginning of experiment section}

In this section, we evaluate the accuracy, consistency, and efficiency of the proposed system on two large-scale multi-session datasets covering high-speed autonomous driving and moderate-speed outdoor robot scenarios. We compare our method against the baseline Fast-LIO2 \cite{xu2022fast} and a recent map merging approach, LAMM \cite{wei2024large}, which also assumes Fast-LIO2 as its SLAM frontend. To isolate the impact of the proposed chained loop closure, we conduct ablation studies by evaluating our system with and without chained loop closure while keeping all other hyperparameters identical.

We assess performance from multiple perspectives:
\begin{itemize}
    \item \textbf{Trajectory accuracy} is evaluated using Absolute Pose Error (APE) over full trajectories.
    \item \textbf{Geometric consistency} is evaluated via inter-session accumulated point cloud registration.
    \item \textbf{Viewpoint alignment} is assessed through regional 3D reconstruction quality.
    \item \textbf{Computational efficiency} is evaluated in terms of runtime and memory usage.
\end{itemize}
% \textbf{Trajectory accuracy} is evaluated using Absolute Pose Error (APE) over full trajectories.
% \textbf{Geometric consistency} is evaluated via inter-session accumulated point cloud registration.
% \textbf{Viewpoint alignment} is assessed through regional 3D reconstruction quality.
% \textbf{Computational efficiency} is evaluated in terms of runtime and memory usage.
All experiments are conducted on the same platform with an Intel i9-12900K CPU and 64\,GB DDR5 5600\,MHz RAM.

\begin{table}[t]
\centering
\caption{Dataset statistics and sensor configurations. Five sessions per dataset. Duration is total time across sessions.}
\label{tab:dataset_stats}
\resizebox{0.7\columnwidth}{!}{
\footnotesize
\begin{tabular}{lcc}
\hline
\textbf{} & \textbf{MARS} & \textbf{NCLT} \\
\hline
\textbf{Duration} & 12 h & 8 h \\
\textbf{Area} & $1500{\times}3000\ m$ & $400{\times}800\ m$ \\
\hline
\textbf{LiDAR} & VLS-128 (10 Hz) & HDL-32E (10 Hz) \\
\textbf{IMU} & 10 Hz & 100 Hz \\
\textbf{GNSS} & 10 Hz & 1 Hz \\
\hline
\end{tabular}
}
\vspace{-6mm}
\end{table}

\subsection{Datasets}

% We evaluate the proposed system on two large-scale multi-session LiDAR datasets: MARS \cite{li2024multiagent} and NCLT \cite{carlevaris2016university}. Both datasets contain repeated traversals of the same environments over extended durations, enabling evaluation of multi-session alignment and long-horizon geometric consistency. Key dataset statistics and sensor configurations are summarized in Table~\ref{tab:dataset_stats}.
% The MARS dataset consists of urban autonomous driving data collected using a high-speed vehicle platform with a dense LiDAR sensor, providing frequent revisits across large spatial extents from varied directions. In contrast, the NCLT dataset contains outdoor data collected using a mobile robot platform operating at moderate speeds, with sparser LiDAR observations and more agile motion characteristics. These differences in sensing density and motion profiles provide complementary large-scale environments for evaluating multi-session SLAM performance.
We evaluate the system on two large-scale multi-session LiDAR datasets: MARS \cite{li2024multiagent} and NCLT \cite{carlevaris2016university}. Both contain repeated traversals over extended durations, enabling evaluation of multi-session alignment and long-horizon geometric consistency. Key dataset statistics and sensor configurations are summarized in Table~\ref{tab:dataset_stats}.
MARS consists of urban autonomous driving data collected with a high-speed vehicle platform and a dense LiDAR sensor, providing frequent revisits across large spatial extents from varied directions. In contrast, NCLT contains outdoor data collected from a mobile robot platform with sparser LiDAR observations and more agile motion. These differences in sensing density and motion profiles provide complementary environments for evaluating large-scale multi-session SLAM performance.

% \begin{figure}[t]
% \centering
% \includegraphics[width=\columnwidth]{fig/dataset_statistics_v0.png}
% \caption{3D Reconstruction Streets Dataset statistics.}
% \label{fig:dataset_statistics}
% \end{figure}

% \begin{figure}[t]
% \centering
% \includegraphics[width=\columnwidth]{fig/Fast-LIO2_map.png}
% \caption{Fast-LIO2 result. No loop closure implemented, resulting in misalignments at revisits.}
% \label{fig:Fast-LIO2_map}
% \end{figure}

% \begin{figure}[t]
% \centering
% \includegraphics[width=\columnwidth]{fig/lamm_result.png}
% \caption{LAMM result. Global optimization struggles to optimize hundreds of loop closures at the same time, resulting in distortions and misalignments.}
% \label{fig:lamm_map}
% \end{figure}

% \begin{table}[t]
% \centering
% \caption{Trajectory Absolute Pose Error in terms of planar translation (XY) and yaw rotation. RMSE is reported in meters for XY and degrees for yaw.}
% \label{tab:ape_xy_yaw}
% \footnotesize
% \begin{tabular}{lcc}
% \hline
% \textbf{Method} & \textbf{XY RMSE (m)} & \textbf{Yaw RMSE (deg)} \\
% \hline
% Fast-LIO2        & 53.8568 & 3.4763 \\
% LAMM            & 23.3530 & 4.1769 \\
% Ours unchained  & 3.9729  & 2.2557 \\
% Ours chained    & \textbf{3.9230}  & \textbf{2.1615} \\
% \hline
% \end{tabular}
% \end{table}

% MARS with z-aligned gt, full xyz rpy APE
% & Fast-LIO2        & 52.62 & 4.71 \\
% & LAMM            & 43.69 & 10.29 \\
% & Ours unchained  & 14.95 & 3.09 \\
% & Ours chained    & \textbf{9.64}  & \textbf{3.45} \\
% \vspace{-20}
\begin{table}[t]
\centering
\caption{Trajectory absolute pose error (APE) evaluations on MARS and NCLT datasets with \colorbox{green!15}{first} and \colorbox{yellow!15}{second} places. 
% \yq{I put an example here using cellcolor to highlight the result, you can consider using it or not.}
}
\label{tab:trajectory_eval}
\resizebox{\columnwidth}{!}{
\begin{tabular}{llcc}
\hline
\textbf{Dataset} & \textbf{Method} & \textbf{T RMSE (m)} & \textbf{R RMSE (deg)} \\
\hline
\multirow{4}{*}{MARS (2D)}
& Fast-LIO2        & 53.857 & 3.476 \\
& LAMM            & 23.353 & 4.177 \\
& Ours unchained  & \cellcolor{yellow!15}4.336  & \cellcolor{yellow!15}2.186 \\
& Ours chained    & \cellcolor{green!15}\textbf{3.923}  & \cellcolor{green!15}\textbf{2.161} \\
\hline
\multirow{4}{*}{MARS (3D)}
& Fast-LIO2        & 52.616 & 4.708 \\
& LAMM            & 43.688 & 10.285 \\
& Ours unchained  & \cellcolor{yellow!15}13.511 &  \cellcolor{yellow!15}3.729 \\
& Ours chained    & \cellcolor{green!15}\textbf{9.640}  & \cellcolor{green!15}\textbf{3.448} \\
\hline
\multirow{4}{*}{NCLT (3D)}
& Fast-LIO2        & 2.279 &  \cellcolor{green!15}\textbf{2.618} \\
& LAMM            & 33.547 & 11.549 \\
& Ours unchained  & \cellcolor{yellow!15}1.473 & 4.876 \\
& Ours chained    & \cellcolor{green!15}\textbf{1.197} & \cellcolor{yellow!15}4.647 \\
\hline
\end{tabular}}
\vspace{-6mm}
\end{table}
% \vspace{-1pc}

\subsection{Trajectory Accuracy: Absolute Pose Error}
\label{subsec:ape}
% We evaluate trajectory accuracy using Absolute Pose Error (APE) in terms of translation and rotation, computed with the \textit{evo} evaluation toolkit. Root Mean Square Error (RMSE) is reported separately for translation and rotation. Experiments are conducted on both the MARS dataset and the NCLT dataset. For MARS, since we observed slight z-drift but very accurate x y coordinates in its ground truth pose, we perform  two versions of evaluation: 2D APE and 3D APE, where 2D involves only XY and Yaw, and 3D includes all of xyz rpy 6 DoF. For NCLT, only 3D 6DoF evaluation is performed. As shown in Table~\ref{tab:trajectory_eval}, the proposed method with chained loop closure achieves the lowest translation RMSE across both datasets, demonstrating improved trajectory alignment and enhanced global geometric consistency relative to ground truth.

% We evaluate trajectory accuracy using Absolute Pose Error (APE) in translation and rotation, computed with the \textit{evo} evaluation toolkit \cite{grupp2017evo}. Root Mean Square Error (RMSE) is reported separately for translation and rotation. Experiments are conducted on both the MARS and NCLT datasets. 
% For the MARS dataset, we observe minor drift in the vertical component of the provided ground truth while the planar components remain highly consistent. Therefore, we report both 2D and 3D APE. The 2D evaluation considers planar translation (XY) and yaw rotation, while the 3D evaluation includes full 6-DoF pose error in translation (XYZ) and rotation (roll, pitch, yaw). For the NCLT dataset, we report only full 6-DoF APE.
We evaluate trajectory accuracy using Absolute Pose Error (APE) in translation and rotation with the \textit{evo} toolkit \cite{grupp2017evo}, reporting RMSE for each. Experiments are conducted on the MARS and NCLT datasets, with trajectory visualizations shown in Fig.~\ref{fig:trajectory}. 
For MARS, minor drift is observed in the vertical component of the ground truth pose, while planar components remain consistent. Therefore, both 2D and 3D APE are reported. The 2D evaluation considers planar translation (XY) and yaw, while the 3D evaluation measures full 6-DoF pose error in translation (XYZ) and rotation (roll, pitch, yaw). For NCLT, only 3D APE is reported.

Since Fast-LIO2 does not produce a unified coordinate frame across sessions, we evaluate each scene separately and report the average RMSE, while LAMM and our method are evaluated on the combined multi-session trajectory. 
As shown in Table~\ref{tab:trajectory_eval}, our method significantly outperforms the baselines on MARS, achieving 5--10$\times$ lower translation errors and 1.5--3$\times$ lower rotation errors. On NCLT, it also achieves lower translation RMSE, showing robustness across different sensing densities and motion patterns. Chained loop closure further reduces both translation and rotation RMSE over the ablation across both datasets, demonstrating its effectiveness in improving trajectory alignment and global geometric consistency.

Notably, Fast-LIO2 reports the lowest rotation RMSE on NCLT. 
This is expected because Fast-LIO2 is evaluated per session and averaged, effectively measuring single-session accuracy. As an odometry-only system coupling IMU propagation with local scan-to-map matching, but without global optimization or loop closure, it prioritizes local accuracy over global alignment.
Consequently, its rotation accuracy benefits from datasets with strong local motion observability. In particular, NCLT provides a high-rate 100\,Hz IMU and slower motion, whereas on the faster MARS platform with a 10\,Hz IMU, Fast-LIO2 exhibits $1.5\times$ higher rotation error than our method. This indicates that our approach maintains robust performance across platforms with reduced dependence on sensor rate and raw motion measurement fidelity.

\begin{figure}[t]
\centering
\includegraphics[width=0.9\columnwidth]{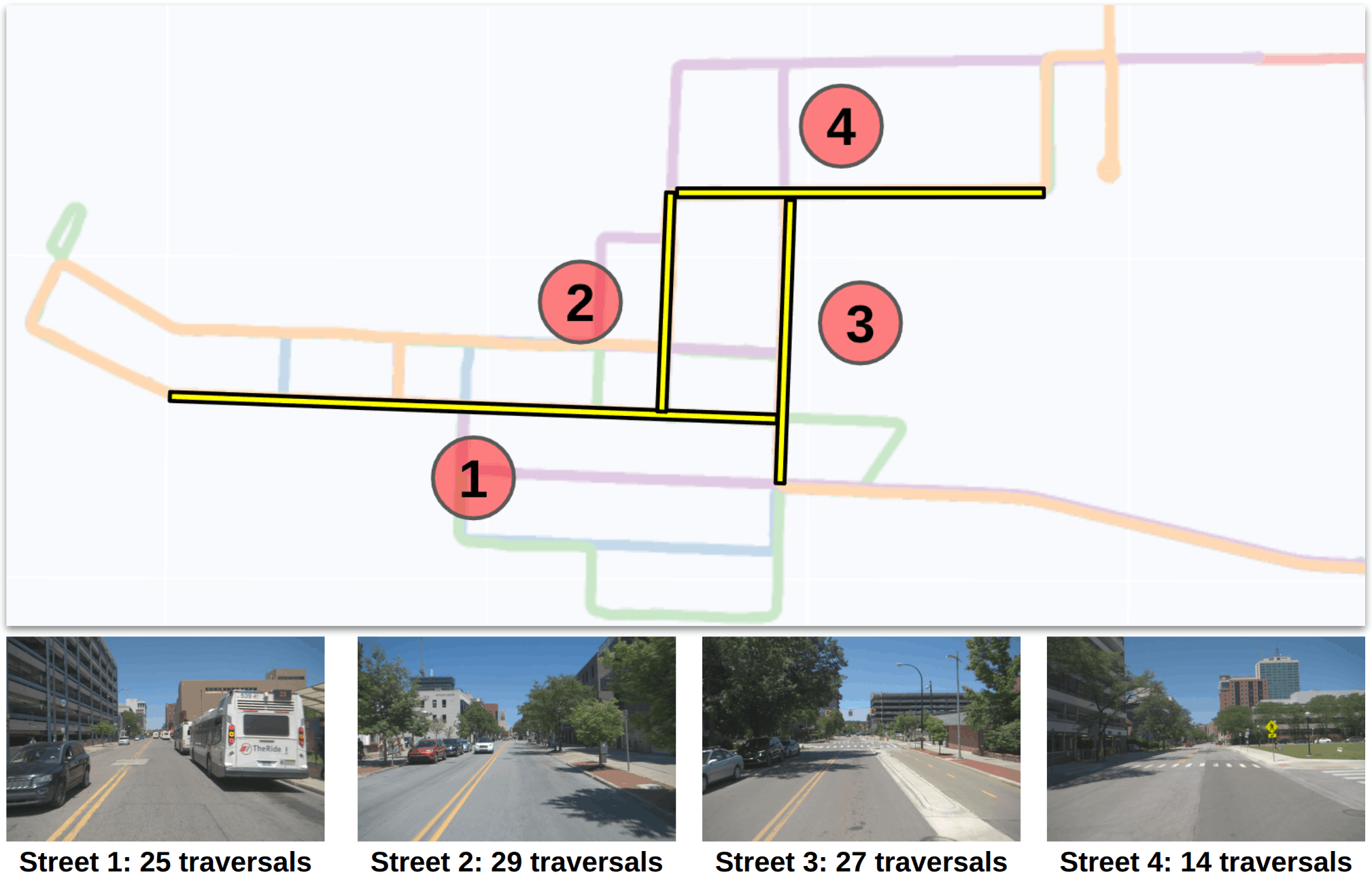}
\caption{Streets with extensive revisits in varied directions selected for point cloud registration and 3D reconstruction.}
\label{fig:streets}
\vspace{-2mm}
\end{figure}

\begin{figure}[t]
\centering
\includegraphics[width=0.9\columnwidth]{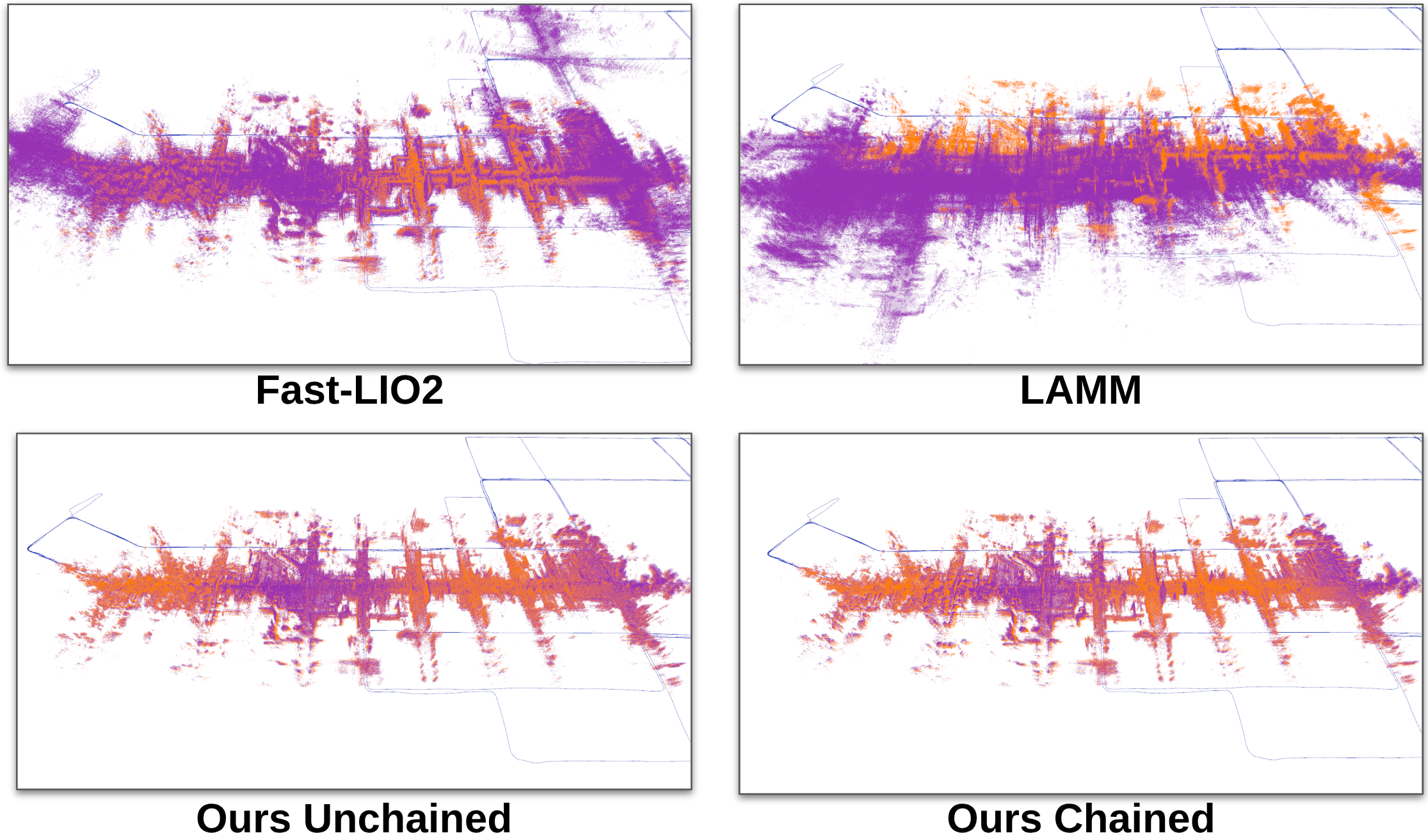}
\caption{Point cloud registration on Street~4 between \textcolor{chained}{\textbf{method output}} and \textcolor{gt}{\textbf{ground truth}}. Close overlap indicates strong geometric consistency. Chained loop closure minimally affects local geometry while improving global trajectory accuracy.}
\label{fig:pcd_registration}
\vspace{-4mm}
\end{figure}

\subsection{Geometric Consistency: Point Cloud Registration}

To evaluate inter-session geometric consistency, we report accumulated point cloud registration RMSE on dense point clouds from the MARS dataset. We focus on four streets with frequent revisits from different directions, with traversal counts shown in Fig.~\ref{fig:streets}. 
Each street is defined by manually selecting its endpoints in GPS coordinates and fitting a centerline; keyframes are assigned to streets based on their Euclidean distance to this line.
For each method, dense point clouds are constructed by accumulating LiDAR scans of the associated keyframes using estimated poses. These are then registered to reference point clouds generated from ground truth poses (z-aligned). The resulting registration RMSE reflects the inter- and intra-session geometric consistency of the estimated trajectory and reconstructed point cloud.

% Registration is performed using RANSAC followed by ICP refinement with identical parameters and fixed random seeds across all methods. We report both RANSAC and ICP-refined RMSE, where lower values indicate better alignment with ground-truth geometry and inter-session consistency. Fast-LIO2 results are computed per scene and averaged due to the lack of unified coordinate frame across sessions, thus reflecting only intra-session consistency.
% As shown in Table~\ref{tab:registration}, our method achieves lower registration errors than the baselines, reducing RANSAC RMSE by approximately $20\%$ compared to Fast-LIO2 and $40\%$ compared to LAMM. 
% The chained variant further yields the lowest ICP-refined errors on most streets, indicating that chained loop closures preserve local geometric consistency (Fig.~\ref{fig:pcd_registration}) while improving global alignment (Tab.~\ref{tab:trajectory_eval}).

Registration is performed using RANSAC followed by ICP refinement, with identical parameters and fixed random seeds across all methods. We report both RANSAC and ICP-refined RMSE, where lower values indicate better alignment with ground-truth geometry and inter-session consistency. Since Fast-LIO2 lacks a unified coordinate frame across sessions, its results are computed per scene and averaged, reflecting only intra-session consistency.
As shown in Table~\ref{tab:registration}, our method achieves lower registration errors than the baselines, reducing RANSAC RMSE by $20\%$ over Fast-LIO2 and $40\%$ over LAMM. 
The chained variant further obtains the lowest ICP-refined errors on most streets, showing that chained loop closures preserve local geometric consistency (Fig.~\ref{fig:pcd_registration}) while improving global alignment (Tab.~\ref{tab:trajectory_eval}).

% \yq{We have table to show results, the sentences here can be more compact. But two important message need to be written down. first, we have about 2x better than existing methods, second, what's the benifit  }

\begin{table}[t]
\centering
\caption{Accumulated Point Cloud Registration RMSE comparison (meters). Values are RANSAC (ICP).}
\label{tab:registration}
\resizebox{\columnwidth}{!}{
\begin{tabular}{lcccc}
\toprule
\textbf{Method} & \textbf{Street 1} & \textbf{Street 2} & \textbf{Street 3} & \textbf{Street 4} \\
\midrule
Fast-LIO2 
& 5.108 (2.551)
& 5.071 (2.425)
& 4.989 (2.513)
& 5.028 (2.473) \\

LAMM 
& 6.650 (2.624)
& 6.652 (2.622)
& 6.510 (2.634)
& 6.739 (2.607) \\

Ours Unchained 
& \textbf{3.764} (\textbf{2.306})
& 4.686 (2.452)
& \textbf{3.701} (2.416)
& 4.211 (2.272) \\

Ours Chained 
& 4.091 (2.400)
& \textbf{3.860} (\textbf{2.120})
& 3.915 (\textbf{2.190})
& \textbf{3.793} (\textbf{2.059}) \\

\bottomrule
\end{tabular}
}
\vspace{-2mm}
\end{table}

\subsection{Viewpoint Alignment: 3D Reconstruction}

To evaluate global geometric consistency, we assess reconstruction quality using test-set PSNR and loss from 3D Gaussian Splatting (3DGS) \cite{kerbl20233d} on four streets from the MARS dataset (Fig.~\ref{fig:streets}). We focus on test-set metrics rather than training PSNR, as misaligned trajectories can still achieve high training PSNR by overfitting individual frames without maintaining consistent viewpoints across neighboring observations. In contrast, accurate pose estimation improves alignment between training and test views, enabling coherent reconstruction of unseen regions. Therefore, higher test PSNR and lower loss indicate improved viewpoint alignment and trajectory consistency across repeated traversals.

As shown in Table~\ref{tab:3dgs_psnr}, our method consistently achieves the best reconstruction quality among the compared approaches, with the chained variant performing strongest overall. While LAMM improves alignment over the odometry-only baseline through global optimization, residual inconsistencies may persist due to local registration errors and trajectory distortion, both of which reduce viewpoint overlap. In contrast, our approach improves cross-session pose alignment and enforces stronger global consistency, enabling more coherent reconstruction across revisited regions. These results demonstrate that improved trajectory consistency directly translates to better viewpoint alignment and reconstruction quality in large-scale multi-session environments.

% \begin{table}[t]
% \centering
% \caption{Average 3DGS test PSNR and loss comparison across streets. Higher PSNR and lower loss are better.}
% \label{tab:3dgs_psnr}
% \resizebox{0.7\columnwidth}{!}{
% \begin{tabular}{lcc}
% \toprule
% \textbf{Method} & \textbf{Avg PSNR} & \textbf{Avg Loss} \\
% \midrule
% Fast-LIO2 & 11.184 & 0.232 \\
% Fast-LIO2+LAMM & 12.216 & 0.205 \\
% Ours Unchained & 12.792 & 0.188 \\
% Ours Chained & \textbf{12.976} & \textbf{0.183} \\
% \bottomrule
% \end{tabular}
% }
% \end{table}

\begin{table}[t]
\centering
\caption{Average 3DGS test PSNR and Loss comparison.}
\label{tab:3dgs_psnr}
\resizebox{0.55\columnwidth}{!}{
\begin{tabular}{lcc}
\toprule
\textbf{Method} & \textbf{PSNR $\uparrow$} & \textbf{Loss $\downarrow$} \\
\midrule
Fast-LIO2 & 11.184 & 0.232 \\
LAMM & 12.444 & 0.198 \\
Ours Unchained & 13.222 & 0.178 \\
Ours Chained & \textbf{13.317} & \textbf{0.174} \\
\bottomrule
\end{tabular}
}
\vspace{-6mm}
\end{table}

% \subsection{Runtime Analysis}
% All systems were run on a PC with Intel i9-12900K with 64GB DDR5 5600MHz RAM.
% For MARS, data come from the 
% \begin{table}[t]
% \centering
% \caption{Runtime and memory usage comparison on the MARS and NCLT datasets. Runtime is measured in seconds per frame and RAM usage in megabytes.}
% \label{tab:runtime_memory}
% \resizebox{\columnwidth}{!}{
% \begin{tabular}{llcc}
% \hline
% \textbf{Dataset} & \textbf{Method} & \textbf{Runtime (s)} & \textbf{RAM (MB)} \\
% \hline
% \multirow{4}{*}{MARS}
% & Fast-LIO2        & 0.02228 & 1065.56 \\
% & LAMM            & 0.04375 & 16821.90 \\
% & Ours unchained  & 0.48433 & 12571.12 \\
% & Ours chained    & 0.81887 & 16744.26 \\
% \hline
% \multirow{4}{*}{NCLT}
% & Fast-LIO2        & 0.00734 & 928.77 \\
% & LAMM            & 0.01865 & 5917.86 \\
% & Ours unchained  & 0.15456 & 10256.43 \\
% & Ours chained    & 0.16086 & 10544.34 \\
% \hline
% \end{tabular}}
% \end{table}

\subsection{Computational Efficiency}

% We evaluate computational performance by measuring runtime and memory usage. Runtime is reported as the average processing time per raw LiDAR frame, separated into frontend and backend. For LAMM, the frontend time is identical to Fast-LIO2 since it directly uses Fast-LIO2 outputs. For our method, the frontend includes processing up to the EKF update, while the backend covers the remaining operations to the end of the frame.
% Frames per second (FPS) is computed as $\frac{1000}{t_{\text{front}} + t_{\text{back}}}$ using the frontend and backend processing times per frame (ms).
% Memory usage is reported as the peak RAM consumption over the entire run. 
% All reported values correspond to the final session, during which all maps from the four previous sessions are loaded and jointly optimized with the current session.

We evaluate computational performance using runtime and peak RAM usage. Runtime is reported as the average processing time per raw LiDAR frame, split into frontend and backend. For LAMM, the frontend time matches Fast-LIO2 since it directly uses Fast-LIO2 outputs. For our method, the frontend includes processing through the EKF update, and the backend includes the remaining per-frame operations. 
Frames per second (FPS) is computed as $1000/(t_{\text{front}} + t_{\text{back}})$, where times are in milliseconds. 
Memory is reported as peak RAM over the full run. 
All values are measured on the final session, where maps from the previous four sessions are loaded and jointly optimized with the current session.

% As shown in Tab.~\ref{tab:runtime_memory}, our system maintains frontend processing time comparable to the Fast-LIO2 baseline, with additional keyframe management and global optimization handled in an asynchronous backend that does not affect frontend odometry propagation. Our memory usage is also slightly lower than LAMM while achieving significantly better alignment. The system operates in real time, running at approximately 2\,Hz on the full 12-hour MARS dataset with 128-channel LiDAR and 13\,Hz on the 8-hour NCLT dataset with 32-channel LiDAR. These results demonstrate that the system improves global consistency while maintaining real-time performance in large-scale, long-duration deployments.

As shown in Tab.~\ref{tab:runtime_memory}, our system keeps frontend processing time comparable to Fast-LIO2, while keyframe management and global optimization run asynchronously in the backend without affecting odometry propagation. Our memory usage is also slightly lower than LAMM while achieving better alignment. The system runs in real time, reaching approximately 2\,Hz on the 12-hour MARS dataset with 128-channel LiDAR and 13\,Hz on the 8-hour NCLT dataset with 32-channel LiDAR. These results show that our method improves global consistency while preserving real-time performance in large-scale, long-duration deployments.

\begin{table}[t]
\centering
\caption{Runtime and memory usage comparison.}
\resizebox{\columnwidth}{!}{
\label{tab:runtime_memory}
\begin{tabular}{llcccc}
\hline
\textbf{Dataset} & \textbf{Method} & \textbf{Frontend (ms)} & \textbf{Backend (ms)} & \textbf{FPS (Hz)} & \textbf{RAM (MB)} \\
\hline
\multirow{4}{*}{MARS}
& Fast-LIO2        & 22.28  & -- & 44.88 & 1065.56 \\
& LAMM            & 22.28  & 21.25 & 22.97 & 16821.90 \\
& Ours unchained  & 29.88 & 402.33 & 2.31 & 12571.12 \\
& Ours chained    & 29.39 & 447.02 & 2.10 & 13668.46 \\
\hline
\multirow{4}{*}{NCLT}
& Fast-LIO2        & 7.34   & -- & 136.17 & 928.77 \\
& LAMM            & 7.34  & 11.31 & 53.61 & 5917.86 \\
& Ours unchained  & 7.79 & 67.50 & 13.28 & 5682.82 \\
& Ours chained    & 7.82 & 70.13 & 12.83 & 5816.96 \\
\hline
\end{tabular}
}
\vspace{-6mm}
\end{table}

% \section{CONCLUSIONS}

% We presented FastLIO-Multi, a real-time LiDAR--inertial SLAM system that enables consistent multi-session and multi-map integration within a unified factor graph framework. The proposed chained loop closure mechanism propagates geometric constraints across connected keyframes, strengthening global consistency and improving robustness over long spatial and temporal horizons without requiring explicit dynamic object removal. Experimental results on large-scale autonomous driving and mobile robot datasets demonstrate improved trajectory accuracy, geometric consistency, and viewpoint alignment compared to existing methods, while maintaining real-time performance on extended multi-session deployments. These results validate the effectiveness of the proposed system for robust long-term mapping and localization in large scale multi-session environments.

% \textbf{Limitations.} \yq{Although our method achieves state-of-the-art multi-session and multi-map reconstruction results, we still have a limitation that the current map merging implementation uses GNSS-proximity based place recognition. While GNSS data is commonly available for large scale dataset, integration with descriptor based place recognitions can further improve generalizability.  (put an objective and non-vital limitation here) }
\section{CONCLUSIONS}

We presented \textbf{\modelname}, a simple and effective LiDAR--inertial SLAM backend for real-time multi-session integration. Its chained loop closure propagates geometric constraints across connected keyframes, improving global consistency and long-term robustness without explicit dynamic object removal. Experiments on large-scale driving and mobile robot datasets show improved trajectory accuracy, geometric consistency, and viewpoint alignment over existing methods, with minimal hyperparameter tuning across platforms and real-time performance across extended multi-session deployments.
These results demonstrate the effectiveness of the proposed system for robust long-term mapping and localization in large-scale multi-session environments.

\textbf{Limitations.}
% Although our method achieves state-of-the-art performance in multi-session trajectory accuracy and global consistency, it still has certain limitations. 
% The current map merging implementation uses GNSS-proximity-based place recognition to identify candidate cross-session correspondences. While GNSS data is commonly available in large-scale datasets, LiDAR-based place recognition methods such as Scan Context \cite{kim2018scan} and STD \cite{yuan2023std} can be integrated to replace the GNSS proximity search stage, resulting in a purely LiDAR–inertial system. This extension would improve generalizability and facilitate deployment in GNSS-denied environments, further enhancing the robustness of the proposed real-time multi-session SLAM system.
% Although our method achieves state-of-the-art multi-session trajectory accuracy and global consistency, map merging currently relies on GNSS proximity to propose cross-session correspondences. GNSS is commonly available in large-scale datasets; however, this cue could be replaced by LiDAR-based place recognition methods such as Scan Context \cite{kim2018scan} or STD \cite{yuan2023std} to enable GNSS-free merging, improving generalizability and supporting deployment in GNSS-denied environments.
While our method achieves state-of-the-art multi-session trajectory accuracy and global consistency, map merging currently relies on GNSS to propose cross-session correspondences. Though GNSS is commonly available in large-scale datasets, this cue could be replaced by LiDAR-based place recognition methods such as Scan Context \cite{kim2018scan} or STD \cite{yuan2023std} for GNSS-free merging, improving generalizability and supporting deployment in GNSS-denied environments.

\section*{Acknowledgement}
The work was supported in part by NSF Grants 2238968 and 2514030, and in part by NYU IT High Performance Computing resources. The authors would also like to thank the anonymous reviewers for their valuable suggestions.

\addtolength{\textheight}{-1cm}   % This command serves to balance the column lengths

\bibliographystyle{IEEEtran}
\bibliography{refs}

\end{document}